\documentclass[]{bytedance_seed}

\usepackage[toc,page,header]{appendix}

\usepackage{minitoc}

\usepackage{amssymb}
\usepackage{adjustbox}
\usepackage{multirow}
\usepackage{makecell}
\usepackage{wrapfig}
\usepackage{float}
\usepackage{hyperref}

\usepackage{mathtools}
\usepackage{pifont}%

\usepackage{algorithm}
\usepackage{algorithmic}

\newcommand{\myPara}[1]{\vspace{.05in}\noindent\textbf{#1.}}

\usepackage{fontawesome5}

\newcommand{\name}{MoDA}

\title{Mixture-of-Depths Attention}

\author[1,2,*]{Lianghui Zhu}
\author[2,\dagger]{Yuxin Fang}
\author[1,2,*]{Bencheng Liao} %
\author[2]{Shijie Wang} %
\author[2]{\\Tianheng Cheng} %
\author[2]{Zilong Huang} %
\author[2]{Chen Chen} %
\author[2]{Lai Wei} %
\author[2]{Yutao Zeng} %
\author[2]{Ya Wang} %
\author[2]{Yi Lin} %
\author[2]{Yu Li} %
\author[1,\text{\faEnvelope[regular]}]{Xinggang Wang}

\affiliation[1]{School of EIC, Huazhong University of Science \& Technology}
\affiliation[2]{ByteDance Seed}

\contribution[\dagger]{Project Lead}
\contribution[\text{\faEnvelope[regular]}]{Corresponding author}

\abstract{
{
  \checkdatafont\sffamily \bfseries Abstract:} 
  Scaling depth is a key driver for large language models (LLMs). 
  Yet, as LLMs become deeper, they often suffer from \emph{signal degradation}: informative features formed in shallow layers are gradually diluted by repeated residual updates, making them harder to recover in deeper layers.
  We introduce \emph{mixture-of-depths attention} (\name), a mechanism that allows each attention head to attend to sequence KV pairs at the current layer and depth KV pairs from preceding layers.
  We further describe a hardware-efficient algorithm for \name{} that resolves non-contiguous memory-access patterns, achieving 97.3\% of FlashAttention-2's efficiency at a sequence length of 64K.
  Experiments on 1.5B-parameter models demonstrate that \name{} consistently outperforms strong baselines. 
  Notably, it improves average perplexity by 0.2 across 10 validation benchmarks and increases average performance by 2.11\% on 10 downstream tasks, with a negligible 3.7\% FLOPs computational overhead.
  We also find that combining \name{} with post-norm yields better performance than using it with pre-norm.
  These results suggest that \name{} is a promising primitive for depth scaling.

  \correspondence{\email{xgwang@hust.edu.cn}}
  \checkdata[Code]{\url{https://github.com/hustvl/MoDA}}
}

\begin{document}

\maketitle

\begin{figure}[htbp]
  \vspace{-4em}
  \centering
  \includegraphics[width=1.\textwidth]{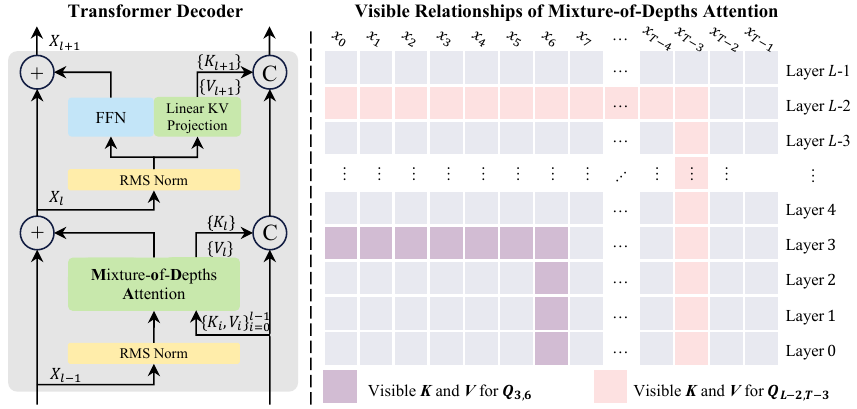}
  \vspace{-5mm}
 \caption{
  We propose mixture-of-depths attention ({\name{}}) to address the modern LLM's information dilution problem in a dynamic and hardware-efficient way.
  Compared with vanilla causal sequence attention, \name{} additionally allows query to attend to depth memories, i.e., depth KV pairs $\{K_i, V_i\}_{i=0}^{l-1}$ at the same query position from preceding layers.
  }
 \label{fig:dsa_pipeline}
\end{figure}
\newcommand{\customfootnote}[2]{%
  \begingroup
    \renewcommand\thefootnote{#1}%
    \footnotetext{#2}%
    \addtocounter{footnote}{-1}%
  \endgroup
}
\customfootnote{*}{This work was done when Lianghui Zhu and Bencheng Liao was interning at
ByteDance Seed.}

  \begin{tcolorbox}[boxsep=0pt, left=10pt, right=10pt, top=5pt, bottom=5pt]
    \setlength{\parindent}{0cm}
    \setlength{\parskip}{0cm}
    \abstractlist\par
    {
      \setlength{\parskip}{0cm}
      \ifdefempty{\checkdatalist}{\vspace*{0cm}}{\checkdatalist\par}
    }
  \end{tcolorbox}
  \tcbset{reset}
  \FloatBarrier

\section{Introduction}
Recent progress in large language models (LLMs)~\cite{team2023gemini,achiam2023gpt4, guo2025deepseek,liu2024deepseek} has been driven by scaling along four major dimensions: context length~\cite{child2019sparse,yuan2025nsa,dai2019transformerxl}, training data~\cite{achiam2023gpt4,team2023gemini}, model width \cite{touvron2023llama,bai2023qwen}, and model depth \cite{wang2024deepnet,chen2026post}.
Although these dimensions remain effective, incremental gains are becoming increasingly costly, motivating interest in complementary architectural scaling strategies.
In current LLM practice, scaling is often realized more through data, context, and especially width, whose optimization behavior and system efficiency are generally easier to realize at scale.
Depth, by contrast, remains comparatively under-exploited despite its strong representational appeal.
In principle, deeper stacks can support richer hierarchical computation.
Yet modern Transformers often fail to convert additional layers into proportional benefits due to the optimization problem~\cite{he2016resnet} and information dilution~\cite{huang2017densenet,pagliardini2024denseformer}.
The resulting question is central to the architecture design: how can a model scale depth while maintaining optimization stability and preventing information dilution?

\begin{figure}[t]
  \centering
  \includegraphics[width=1\textwidth]{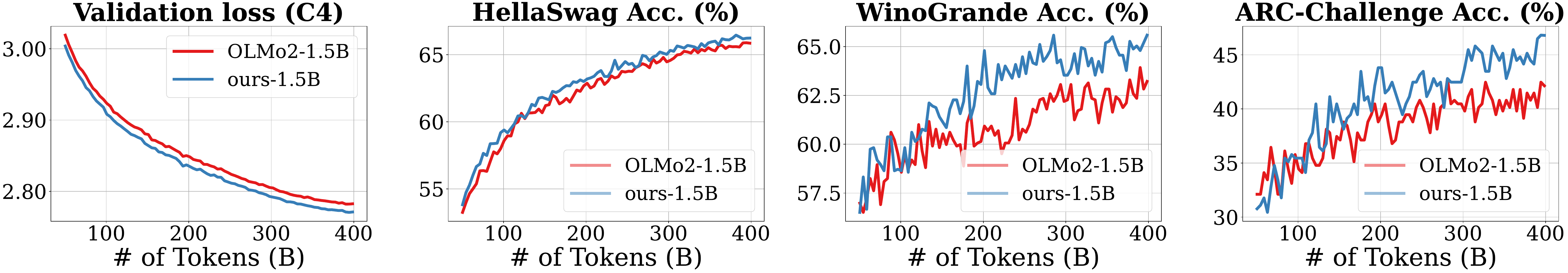}
  \vspace{-6mm}
 \caption{
  Comparing \name{} and strong open-sourced baseline, i.e., OLMo2~\cite{olmo2024olmo2}, with validation loss and downstream performance under the 1.5B-parameter setting.
  Models using \name{} achieve lower C4~\cite{raffel2020c4} validation loss and better downstream performance, i.e., HellaSwag~\cite{zellers2019hellaswag}, WinoGrande~\cite{sakaguchi2021winogrande}, and ARC-Challenge~\cite{clark2018arc}, than OLMo2.
  }
 \label{fig:overall_results}
 \vspace{-2mm}
\end{figure}

The standard residual pathway (ResNet-style) improves optimization stability in deep networks \citep{he2016resnet}, but it still compresses depth history into a single hidden-state trajectory, leaving information dilution largely unresolved.
Many methods~\cite{zhu2024hc,xie2025mhc,li2025vwn} have been tried to address this problem by upgrading the residual connection.
Dense cross-layer connections (DenseNet-style) preserve richer layer-wise history and thus mitigate information dilution \citep{huang2017densenet,pagliardini2024denseformer,chen2017dpn}, but their parameter growth is substantial at LLM scale, which has limited their adoption as a mainstream architecture.
The success of attention~\cite{vaswani2017attention} in sequence modeling suggests a broader principle: data-dependent dynamic mixing can preserve and retrieve historical information more effectively than fixed-pattern aggregation.
This motivates extending the same principle from sequence modeling to depth modeling, i.e., enabling each layer to adaptively read useful states from earlier layers.
Adaptive cross-layer retrieval is therefore promising, yet practical designs still require a better balance among expressivity, efficiency, and hardware friendliness.

In this work, we introduce \emph{mixture-of-depths attention} (\name{}), a unified attention mechanism in which each head jointly attends to sequence KV of the current layer and depth KV from all preceding layers.
Methodologically, we analyze Transformer stacking through a ``read, operate, write'' lens, comparing depth residual, depth dense, and depth attention in a common design space.
\name{} occupies an efficient point that preserves data-dependent depth retrieval without dense cross-layer overhead.

To make \name{} practical at scale, 
we develop a hardware-aware implementation~\cite{dao2022flashattention,dao2023flashattentionv2,yang2024fla} that fuses sequence and depth attention in one forward pass with shared online-softmax states.
Besides, the proposed chunk-aware depth-KV layout and group-aware indexing significantly improve memory access efficiency.
This fused kernel reaches 97.3\% of FlashAttention-2 efficiency at 64K sequence length, showing that depth-aware aggregating can be integrated without sacrificing modern GPU efficiency.

We validate \name{} on decoder-only language models trained with the 400B-token OLMo2 recipe \citep{olmo2024olmo2} at 700M and 1.5B scales.
In our main 1.5B setting, \name{} improves average perplexity by 0.2 across 10 validation benchmarks and increases average downstream performance by 2.11\% on 10 tasks.
We also find that combining \name{} with post-norm yields better performance than using it with pre-norm.
Additional analyzes, i.e., model-size scaling, attention visualization, and layer-number studies, show robust gains and reduced attention-sink~\cite{xiao2023attnsink} behavior via better probability allocation to informative sequence and depth KV.

The contributions of this paper are summarized as:
\begin{itemize}
\item We propose \name{}, a unified attention formulation for dynamic mixtures of sequence and depth, which improves the aggregation of depth-wise information and addresses the information dilution problem of modern LLMs in a data-dependent way.
\item We present a hardware-efficient fused algorithm that makes \name{} practical for long-context LLM training. 
It reaches 97.3\% of FlashAttention-2 efficiency at 64K sequence length with numerical precision within the allowed range.
\item We provide extensive empirical evidence and comprehensive ablations that \name{} consistently and substantially outperforms the strong open-source baseline, OLMo2, across large-scale corpora at multiple model scales, validating each design choice and establishing \name{} as a reliable foundation for depth scaling in LLMs.
\end{itemize}

\section{Mixture-of-Depths Attention}
\subsection{{Preliminary}}
Most modern large language models are built on the Transformer architecture \citep{vaswani2017attention}, where self-attention is the primary token-mixing operator. 
Given a sequence of $T$ tokens $X=(x_1, x_2, \ldots, x_T)\in\mathbb{R}^{T\times D}$ (with hidden dimension $D$), self-attention first projects tokens into queries ($Q$), keys ($K$), and values ($V$) via trainable matrices $W_Q\in\mathbb{R}^{D\times (H_qd)}$ and $W_K, W_V\in\mathbb{R}^{D\times (H_kd)}$.
Under grouped query attention (GQA)~\cite{ainslie2023gqa}, $H_q=G H_k$, $H_k=H_v$, and $D=H_qd$:
\begin{equation}
    Q = X W_Q,\ \ \ \ \ \ \ \ \ \  K = X W_K,\ \ \ \ \ \ \ \ \ V = X W_V,
\end{equation}
where $Q\in\mathbb{R}^{T\times (H_qd)}$ and $K,V\in\mathbb{R}^{T\times (H_kd)}$.
The attention operator computes pairwise similarity between queries and keys, applies a softmax to obtain per-head attention weights $A_h\in \mathbb{R}^{T\times T}$, and returns a weighted sum of values:
\begin{equation}
    \mathrm{Attention}(Q, K, V) = \mathrm{Concat}_{h=1}^{H_q}\left(\mathrm{softmax}\left(\frac{Q_h K_{\phi(h)}^T}{\sqrt{d}} + \mathcal{M}\right)V_{\phi(h)}\right)
\end{equation}
where $Q_h\in\mathbb{R}^{T\times d}$, $K_j,V_j\in\mathbb{R}^{T\times d}$, and $\phi(h)=\lceil h/G\rceil$ maps each query head to its shared key-value head.
Here, $\mathcal{M}\in \mathbb{R}^{T\times T}$ is an additive attention mask.
For causal attention, $\mathcal{M}_{ij}=0$ if $j\leq i$ and $\mathcal{M}_{ij}=-\infty$ otherwise.
For full attention, $\mathcal{M}$ is all zeros.

\begin{figure}[htbp]
    \centering
    \includegraphics[width=1\textwidth]{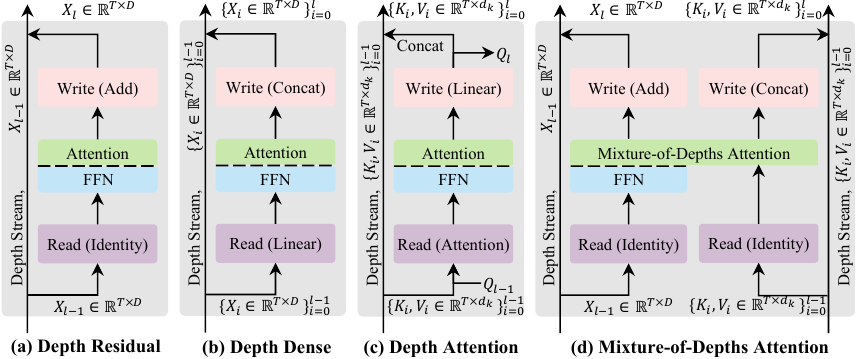}
    \vspace{-4mm}
   \caption{
    Conceptual comparison of mechanisms that utilize the depth stream.
    (a) \textbf{Depth Residual}~\cite{he2016resnet} is the standard residual connection along depth: it reads the current representation and writes back by addition.
    (b) \textbf{Depth Dense}~\cite{huang2017densenet, pagliardini2024denseformer} reads a set of historical representations and linearly projects them back to width $D$; it writes back by concatenation along depth, preserving all intermediate states.
    (c) We introduce \textbf{Depth Attention} as an intermediate formulation, which uses attention to read historical depth KV pairs in a data-dependent way. It writes back by concatenating the current layer's keys and values along depth.
    (d) We propose the upgraded version of Depth Attention, i.e., \textbf{Mixture-of-Depths Attention} (\name), which combines depth attention with standard sequence attention. It writes both the current layer's output and its KV pairs to depth streams for subsequent layers.
    }
   \label{fig:dsa_variants}
\end{figure}

\subsection{Stacking Transformers Along the Depth Stream}

Deep neural networks have enabled breakthroughs across domains, especially after the introduction of residual connections~\cite{he2016resnet}. 
Scaling studies~\cite{kaplan2020scalinglaw,hoffmann2022chinchilla,hestness2017deepscaling} further show that increasing depth can substantially improve performance~\cite{simonyan2014vdcn,szegedy2015godeep}. 
This motivates a natural question:

\textit{Is the residual connection the optimal mechanism for propagating information through depth stream?}

Along the depth stream, we can view a Transformer block as a three-step procedure: \textbf{read, operate, and write}. 
We use this lens to describe different mechanisms for stacking Transformer blocks.
For clarity, the first two mechanisms (Depth Residual~\cite{he2016resnet}, Depth Dense~\cite{huang2017densenet, pagliardini2024denseformer}) are reference designs used to define the depth-stream design space.
We introduce Depth Attention as an intermediate formulation and conceptual bridge.
Our major technical contribution in this section starts from \textbf{Mixture-of-Depths Attention} (\name{}), which unifies sequence and depth retrieval in one unified softmax operator.

\myPara{Depth Residual} 
In depth residual connections~\cite{he2016resnet,srivastava2015highway}, the ``read'' step is identity and the ``write'' step is add.
The ``operate'' step is the token-mixing operator, i.e., attention, or the feed-forward network (FFN), denoted by $\mathcal{F}(\cdot)$.
As shown in Fig.~\ref{fig:dsa_variants}(a), the structure of depth residual can be formulated as follows:
\begin{equation}
  X_{l}=X_0+\sum^{l-1}_{i=1}\mathcal{F}(X_i, \mathcal{W}_{i}),
\end{equation}
where $\mathcal{W}_{i}$ is the set of trainable weight matrices for the $i$-th layer.

This formulation alleviates vanishing gradients and enables training deep networks. 
However, the depth stream is continuously \emph{compressed} into a fixed-size tensor $X_l\in \mathbb{R}^{T\times D}$ via repeated superposition, which dilutes salient features and leads to signal degradation.

\myPara{Depth Dense}
To mitigate signal degradation, depth-dense methods~\cite{huang2017densenet,pagliardini2024denseformer} connect all layers along the depth stream.
At the ``read'' step, they form the input to layer $l$ by linearly projecting the set of preceding representations $\{X_i\in\mathbb{R}^{T\times D}\}_{i=0}^{l-1}$ back to shape $T\times D$.
At the ``write'' step, the layer output is concatenated with the historical set along depth.
As shown in Fig.~\ref{fig:dsa_variants}(b), the structure of depth dense can be formulated as follows:
\begin{equation}
  \{X_{i}\}_{i=0}^{l}=\{X_0, \mathcal{F}(\{X_0\}, \mathcal{W}_{1}), \mathcal{F}(\{X_0, X_1\}, \mathcal{W}_{2}), \cdots ,\mathcal{F}(\{{X_i}\}_{i=0}^{l-1}, \mathcal{W}_{l})\},
\end{equation}
where $\mathcal{W}_{i}$ is the set of trainable weight matrices for the $i$-th layer.

Depth-dense connections propagate information through depth \emph{losslessly}, because concatenation does not compress the historical set.
However, they incur high cost and enforce a fixed connectivity pattern: the computation grows as $O(TL^2D^2)$ in dominant terms, which is prohibitive for large models.

\myPara{Depth Attention}
To reduce cost while retaining adaptive connectivity, we propose \emph{depth attention} that reads historical depth information using attention in a data-dependent way, as illustrated in Fig.~\ref{fig:dsa_variants}(c).
At the ``read'' step, in the GQA-group view ($H_kd=D/G$), we denote one query-group representation by $Q_{l-1}\in\mathbb{R}^{T\times \frac{D}{G}}$ and the corresponding historical key-value sets by $\{K_i\in\mathbb{R}^{T\times \frac{D}{G}}\}_{i=0}^{l-1}$ and $\{V_i\in\mathbb{R}^{T\times \frac{D}{G}}\}_{i=0}^{l-1}$.
The resulting input $X_{l}^{\text{in}}$ is then fed into the ``operate'' step:
\begin{equation}
  X_{l}^{\text{in}} = \mathrm{Attention}(Q_{l-1}, \{{K_i}\}_{i=0}^{l-1}, \{{V_i}\}_{i=0}^{l-1}),
\end{equation}
where attention is performed along the \emph{depth} dimension: for token $t$, the query $Q_{l-1,t}$ attends only to the depth keys and values $\{K_{i,t}, V_{i,t}\}_{i=0}^{l-1}$ from the \emph{same} token position across layers.
After the ``operate'' step, the current layer output $X_{l}^{\text{out}}$ is fed to the ``write'' step, which produces new query/key/value projections:
\begin{equation}
    Q_{l} = X_{l}^{\text{out}} W_{Q, l}^\text{W},\ \ \ \ \ \ \ \ \ \  K_l = X_{l}^{\text{out}} W_{K, l}^\text{W},\ \ \ \ \ \ \ \ \ V_l = X_{l}^{\text{out}} W_{V, l}^\text{W},
\end{equation}
where $W_{Q, l}^\text{W}, W_{K, l}^\text{W}, W_{V, l}^\text{W}\in\mathbb{R}^{D\times \frac{D}{G}}$ are trainable matrices for the layer-$l$ ``write'' operation, and $Q_l,K_l,V_l\in\mathbb{R}^{T\times \frac{D}{G}}$ denote per-group projections.
We concatenate $K_l$ and $V_l$ along depth for future reads, while $Q_l$ is passed forward to the next layer.

Compared with depth-dense connections, depth attention reads historical information adaptively with much lower cost.
Its computation scales as $O(TL^2D)$, which is a factor of $\frac{1}{D}$ smaller than depth dense.

\myPara{Mixture-of-Depths Attention}
Building upon the Depth Attention, we now propose \emph{mixture-of-depths attention} (\name{}).
\name{} adds depth-level information to standard sequence-level attention and fuses these operations into a single operator.
As illustrated in Fig.~\ref{fig:dsa_pipeline} and Fig.~\ref{fig:dsa_variants}(d), \name{} reads the current hidden state $X_{l-1}$ and the historical depth KV stream $\{(K_i, V_i)\}_{i=0}^{l-1}$.
During the ``operate'' step, we apply \name{} to enable each token to attend to both the sequence-level keys and values and its own historical depth-wise keys and values, with all attention scores normalized jointly under a single softmax function.
The implementation detail of \name{} is presented in Alg.~\ref{alg:moda_forward}.
At the ``write'' step, for the attention layer, we append the current layer's key-value pair to the depth stream so that subsequent layers can access them.
For the FFN layer, we obtain its corresponding key-value pair via a light-weight KV projection.

Overall, \name{} provides an efficient, data-dependent mechanism for exploiting depth history with substantially lower overhead than dense cross-layer connectivity.
Furthermore, aggregating the sequence and depth information in one softmax operation provides a uniform representation space.

\begin{table}[t]
  \caption{Asymptotic complexity of depth-stream mechanisms. Here, $T$ is the sequence length, $D$ is the model width, $G$ is the group size of Group Query Attention (GQA)~\cite{ainslie2023gqa}, $H_k$ is the number of key heads (equal to value heads $H_v$), $H_q$ is the number of query heads (=$GH_k$), $d$ is the head dimension, and $L$ is the number of layers. We report dominant terms and omit constant factors.}
  \label{tab:complexity}
  \begin{center}
  \footnotesize
  \setlength{\tabcolsep}{4pt}
  \begin{tabular}{l c c c}
  \toprule
  Methods & Depth Dense & Depth Attention & Mixture-of-Depths Attention \\
  \midrule
  Is data-dependent? & \ding{56} & \ding{52} & \ding{52} \\
  Is unified softmax? & \ding{56} & \ding{56} & \ding{52} \\
  \midrule
  Parameters & \makecell{$\frac{1}{2}G^2L^2H_k^2d^2+\frac{1}{2}G^2LH_k^2d^2$\\\textcolor{blue}{$O(L^2D^2)$}} & \makecell{$G^2LH_k^2d^2+2GLH_k^2d^2$\\\textcolor{blue}{$O(LD^2)$}} & \makecell{$2GLH_k^2d^2$\\\textcolor{blue}{$O(L\frac{D^2}{G})$}} \\
  \midrule
  Decoding Cache & \makecell{$GLH_kd$\\\textcolor{blue}{$O(LD)$}} & \makecell{$2LH_kd$\\\textcolor{blue}{$O(L\frac{D}{G})$}} & \makecell{$2LH_kd$\\\textcolor{blue}{$O(L\frac{D}{G})$}}\\ %
  \midrule
  Prefilling Cache & \makecell{$GTLH_kd$\\\textcolor{blue}{$O(TLD)$}} & \makecell{$2TLH_kd$\\\textcolor{blue}{$O(TL\frac{D}{G})$}} & \makecell{$2TLH_kd$\\\textcolor{blue}{$O(TL\frac{D}{G})$}}\\ %
  \midrule
  Decoding FLOPs & \makecell{$G^2LH_k^2d^2+G^2L^2H_k^2d^2$\\\textcolor{blue}{$O(L^2D^2)$}} & \makecell{$2GL^2H_kd+2GLH_kd$\\\textcolor{blue}{$O(L^2D)$}} & \makecell{$2GL^2H_kd+2GLH_kd$\\\textcolor{blue}{$O(L^2D)$}} \\ %
  \midrule
  Prefilling FLOPs & \makecell{$G^2TLH_k^2d^2+G^2TL^2H_k^2d^2$\\\textcolor{blue}{$O(TL^2D^2)$}} & \makecell{$2GTL^2H_kd+2GTLH_kd$\\\textcolor{blue}{$O(TL^2D)$}} & \makecell{$2GTL^2H_kd+2GTLH_kd$\\\textcolor{blue}{$O(TL^2D)$}} \\ %
  \bottomrule
  \end{tabular}
  \end{center}
  \end{table}

\myPara{Complexity analysis}
Complexity analysis is critical for modern LLM design, we also present the detailed complexity analysis among depth-aware designs, e.g., depth dense, depth attention, and \name{}.
Table~\ref{tab:complexity} reports complete complexity and dominant asymptotic terms, where $T$ is sequence length, $D$ is model width, $L$ is the number of layers, head dimension $d$, and $G$ is the GQA group size. Notably, $H_q = G H_k$.

From Table~\ref{tab:complexity}, Depth Dense is dominated by quadratic depth growth. 
Its parameter term is $O(L^2D^2)$, decoding cache is $O(LD)$, and both decoding and prefilling FLOPs contain quadratic-depth and quadratic-width terms, i.e., $O(L^2D^2)$ and $O(TL^2D^2)$.
The proposed Depth Attention is a data-dependent method, which removes the dominant quadratic-width projection accumulation across depth, reducing parameters to $O(LD^2)$.
It also lowers cache to $O(LD/G)$ and compute to $O(L^2D)$ and $O(TL^2D)$ for decoding and prefilling, respectively.
Compared with Depth Attention, \name{} keeps the same favorable FLOPs order and cache order, but further reduces parameter complexity from $O(LD^2)$ to $O(LD^2/G)$.
The key reason is that \name{} reuses the query projection from sequence attention, so no extra depth-query projection is introduced.
Especially in GQA settings, only grouped depth key/value projections are needed.
This makes \name{} the most parameter-efficient option in Table~\ref{tab:complexity}, while preserving linear-in-width compute behavior and low-cache scaling.

Overall, Table~\ref{tab:complexity} shows that \name{} keeps the data-dependent behavior of attention while avoiding the dominant quadratic-depth parameter growth overhead of dense cross-layer connections.
\name{} aggregates sequence and depth information with a unified softmax operator, which provides better representation and efficiency in practice, especially in regimes with large $L$ and long $T$.

\section{Hardware-aware efficient \name}
\label{sec:hardware_efficient}
Na\"ively PyTorch-implemented~\cite{paszke2019pytorch} \name{} requires non-contiguous reads of historical depth states, which degrades GPU utilization.
We develop a hardware-aware implementation that reorganizes depth-stream tensors to enable contiguous memory access and fused computation.

\begin{algorithm}[t]
\caption{\name{}: Hardware-aware Forward Pass}
\label{alg:moda_forward}
\begin{algorithmic}[1]
\STATE \textbf{Input:} $\mathbf{Q}\in\mathbb{R}^{T_q\times (H_kd)}$, $\mathbf{K},\mathbf{V}\in\mathbb{R}^{T_{kv}\times (H_kd)}$,  $\mathbf{K}^{\text{depth}},\mathbf{V}^{\text{depth}}\in\mathbb{R}^{(T_{kv}L)\times (H_kd)}$, group number $G$
\STATE \textbf{Output:} $\mathbf{O}\in\mathbb{R}^{T_q\times (H_kd)}$%
\STATE Partition $\mathbf{Q},\mathbf{K},\mathbf{V}$ into hardware-friendly query/key/value blocks
\STATE Ensure each query block aligns with $G$ for correct base-time mapping
\FOR{each query block index $b_q$}
    \STATE Load $\mathbf{Q}_{[b_q]}$ from HBM to SRAM (on chip)
    \STATE Initialize on-chip states: $m\leftarrow-\infty,\ acc\leftarrow 0,\ o\leftarrow 0$
    \STATE For each query row index $i_q$ in block $b_q$, compute base-time: $t_{base}(i_q)=\lfloor i_q/G\rfloor$
    \STATE Let $t_{base}^{start}=\min_{i_q\in b_q} t_{base}(i_q)$ and $t_{base}^{last}=\max_{i_q\in b_q} t_{base}(i_q)$
    \STATE Define $t_{base}^{end}=t_{base}^{last}+1$ as the exclusive upper bound
    \FOR{sequence key block $b_s$ with $b_s<t_{base}^{start}$}
        \STATE Load $(\mathbf{K}_{[b_s]},\mathbf{V}_{[b_s]})$ from HBM to SRAM
        \STATE On chip, compute $S=\frac{\mathbf{Q}_{[b_q]}\mathbf{K}_{[b_s]}^\top}{\sqrt{d}}$%
        \STATE On chip, calculate $\textsc{OnlineSoftmaxUpdate}(m,acc,o,S,\mathbf{V}_{[b_s]})$, i.e., the next two lines:
        \STATE On chip, $m'=\max(m,\max S)$, $acc'=acc\cdot 2^{m-m'}+\textstyle\sum 2^{S-m'},\ 
        o'=o\cdot 2^{m-m'}+\textstyle\sum 2^{S-m'}\mathbf{V}_{[b_s]}$
        \STATE On chip, update $(m,acc,o)\leftarrow(m',acc',o')$ %
    \ENDFOR
    \FOR{sequence key block $b_s$ with $t_{base}^{start}\le b_s<t_{base}^{end}$}
        \STATE Load $(\mathbf{K}_{[b_s]},\mathbf{V}_{[b_s]})$ from HBM to SRAM
        \STATE Denote by $i_k$ the sequence-key index in the current key block.
        \STATE On chip, compute $S=\frac{\mathbf{Q}_{[b_q]}\mathbf{K}_{[b_s]}^\top}{\sqrt{d}}$ and apply grouped causal mask $(\lfloor i_q/G\rfloor\ge i_k)$
        \STATE On chip, update $(m,acc,o)\leftarrow\textsc{OnlineSoftmaxUpdate}(m,acc,o,S,\mathbf{V}_{[b_s]})$
    \ENDFOR
    \FOR{depth block index $b_d$ with $t_{base}^{start}L\le b_d<t_{base}^{end}L$}
        \STATE Load $(\mathbf{K}^{\text{depth}}_{[b_d]},\mathbf{V}^{\text{depth}}_{[b_d]})$ from HBM to SRAM
        \STATE Denote by $j_d$ the flattened depth-column index in the current block, i.e., $j_d\in\mathrm{cols}(b_d)$.
        \STATE On chip, compute $S_d=\frac{\mathbf{Q}_{[b_q]}(\mathbf{K}^{\text{depth}}_{[b_d]})^\top}{\sqrt{d}}$ and apply $mask(i_q,j_d)=\mathbf{1}[\lfloor i_q/G\rfloor=\lfloor j_d/L\rfloor]$
        \STATE On chip, update $(m,acc,o)\leftarrow\textsc{OnlineSoftmaxUpdate}(m,acc,o,S_d,\mathbf{V}^{\text{depth}}_{[b_d]})$
    \ENDFOR
    \STATE On chip, normalize $o\leftarrow o/acc$
    \STATE Store output block $\mathbf{O}_{[b_q]}$ from SRAM to HBM
\ENDFOR
\STATE \textbf{return} $\mathbf{O}$%
\end{algorithmic}
\end{algorithm}

\subsection{Preliminary}
Modern GPUs are optimized for throughput-oriented, large-scale data-parallel workloads, where the same operation is applied to many elements in parallel~\cite{yang2023gla,yang2024deltanet,yang2024gdn,dao2022flashattention,dao2023flashattentionv2}.
Therefore, efficient attention kernels should be organized to expose regular, massively parallel computation rather than irregular per-element control flow.

\myPara{Streaming multiprocessors (SMs)}
An NVIDIA GPU is composed of many SMs, which are the basic on-chip units for parallel execution and resource management.
High utilization requires enough independent blocks to keep many SMs active.
In large language model (LLM) training with long-context sequences and relatively small batch sizes, 
parallelization along the temporal dimension is especially important.

\myPara{Compute units: CUDA cores vs. Tensor Cores}
Within each SM, instructions are dispatched to different execution units.
CUDA cores support general arithmetic instructions, while Tensor Cores provide much higher throughput for structured matrix multiply-accumulate operations.
As a result, practical high-performance kernels should maximize regular matmul-style computation to better exploit Tensor Cores.

\myPara{Memory hierarchy: HBM and on-chip SRAM}
End-to-end performance is jointly determined by compute throughput and data movement.
HBM offers large capacity but higher access latency, whereas on-chip SRAM structures, i.e., registers, shared memory, and cache, are much faster but limited in size.
Hence, a key design principle is to improve tiling and data reuse so that hot data stays on chip and HBM traffic is minimized.

These principles directly motivate our hardware-aware \name{} design.
We reorganize depth KV layout and fuse computation to reduce non-contiguous memory access and improve effective compute utilization.

\begin{figure}[t]
    \centering
    \includegraphics[width=1\textwidth]{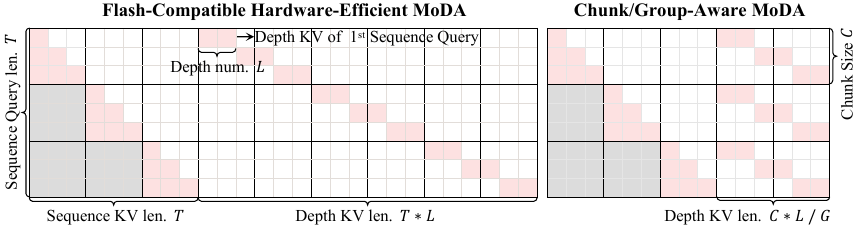}
    \vspace{-4mm}
   \caption{
    Hardware view of \name{} depth-cache access.
    \textbf{Left:} flash-compatible hardware-efficient \name{} keeps a depth KV cache of length $T\times L$ for each sequence, so each query potentially scans a long concatenated depth KV.
    \textbf{Right:} chunk-aware \name{} groups queries by chunk size $C$ and reorganizes depth KV by chunk, reducing the effective depth span from $T\times L$ to $(C\times L)/G$ per chunk,
    where $G$ is the GQA group number.
    This layout improves depth KV calculation efficiency and reduces memory access overhead.
    }
   \label{fig:moda_hardware}
\end{figure}

\subsection{Hardware-aware Considerations for \name}

\myPara{Flash-Compatible depth KV layout}
Na\"ively implementing depth attention with explicit PyTorch for-loops over historical depth KV is typically slow on GPUs,
because it induces irregular gather-like memory access and under-utilizes tensor-core-friendly block compute.
Our first step is a flash-compatible depth-KV layout that flattens the depth cache along a single axis of length $T\times L$.
Thus for each sequence position $t$, its $L$ depth states are stored contiguously.
In this way, each query only needs to map to its corresponding depth range $[tL,(t+1)L)$ to access the correct depth KV slice.
This turns depth lookup into contiguous block reads and makes the depth phase compatible with FlashAttention-style kernels.
Although this flattened formulation is substantially faster than explicit PyTorch for-loops over historical depth KV,
it still introduces a compute-efficiency issue in the depth phase.
In the depth-score matrix $\mathbf{S}^{\text{depth}}\in\mathbb{R}^{T\times (TL)}$, only a block-diagonal region is valid.
Specifically, for query row $i_q$, only depth-column indices $j_d\in[i_qL,(i_q+1)L)$ are needed, while the remaining entries are masked.
We define this ratio as depth utilization, i.e., if computed densely over the full $T\times (TL)$ matrix, 
the depth utilization is $\eta_{\text{depth}}=\frac{T\cdot L}{T\cdot (T\cdot L)}=\frac{1}{T}$.

\myPara{Chunk-aware depth KV layout}
As illustrated in Fig.~\ref{fig:moda_hardware}, 
flash-compatible depth KV layout forces each query block to traverse a long vectorized concatenated depth axis of length $T\times L$, which is unfavorable for depth utilization.
We therefore reorganize depth KV in a chunk-aware manner, i.e., queries are divided into chunks, and each chunk only accesses the corresponding depth-KV span for its covered range.
From a chunk-aware perspective, a query chunk of length $C$ is paired with a local depth-KV region of size $C\times L$, constructed by concatenating the $L$ depth states of the covered $C$ sequence positions.
The kernel therefore computes chunked depth attention over this packed $C\times L$ region, rather than scanning the global $T\times L$ depth axis for every chunk.
This local layout substantially reduces unnecessary HBM traffic from masked, out-of-range depth entries and improves depth utilization to $\eta_{\text{depth}}=\frac{T\cdot L}{T\cdot (C\cdot L)}=\frac{1}{C}$.

\myPara{Group-aware depth KV calculation}
Our key observation is that, under the mapping $T_q = G\,T_{kv}$, 
$G$ adjacent query rows share the same base-time index $\lfloor i_q/G\rfloor$ and can therefore reuse the same depth KV blocks.
Based on this, we design a group-aware depth-KV computation, i.e., for a query chunk of length $C$, only $C/G$ base-time rows are unique, so the required depth span is $(C/G)\times L$ rather than $C\times L$.
Under the fused block-matmul and mask execution, this increases effective depth utilization to $\frac{G\times L}{C\times L}=\frac{G}{C}$.
The same base-time mapping is used consistently in both masks, i.e., $\lfloor i_q/G\rfloor \ge i_k$ for sequence causality and $\lfloor i_q/G\rfloor = \lfloor j_d/L\rfloor$ for depth matching. %
Notably, $i_k$ is the sequence-key index, while $j_d$ is the flattened depth-column index.
In practice, we also align query-block boundaries with $G$, i.e., make block size divisible by $G$, 
to avoid cross-group boundary handling inside one tile and simplify vectorized execution.

\subsection{Hardware-Efficient \name{} Implementation}

\myPara{Preparation}
Algorithm~\ref{alg:moda_forward} follows the group-aware mapping $T_q=G\,T_{kv}$.
The inputs are query $\mathbf{Q}\in\mathbb{R}^{T_q\times (H_kd)}$, sequence key/value
$\mathbf{K},\mathbf{V}\in\mathbb{R}^{T_{kv}\times (H_kd)}$, and depth key/value
$\mathbf{K}^{\text{depth}},\mathbf{V}^{\text{depth}}\in\mathbb{R}^{(T_{kv}L)\times (H_kd)}$,
with output $\mathbf{O}\in\mathbb{R}^{T_q\times (H_kd)}$ and $H_kd=D/G$.
For notation clarity, $b_q,b_s,b_d$ denote block indices, while $i_q,i_k,j_d$ denote element indices inside a block.

Before entering the main loops, all tensors are tiled into hardware-friendly blocks, and each query block is aligned to $G$.
For each query block $b_q$, we load $\mathbf{Q}_{[b_q]}$ from HBM to SRAM and initialize on-chip online-softmax states $(m,acc,o)$,
where $m$ is the running maximum logit, $acc$ is the running softmax normalizer, and $o$ is the running unnormalized output accumulator.
For each query row index $i_q$ in $b_q$, we compute its base-time index
$t_{base}(i_q)=\lfloor i_q/G\rfloor$, and define
$t_{base}^{start}=\min_{i_q\in b_q}t_{base}(i_q)$ and
$t_{base}^{end}=\max_{i_q\in b_q}t_{base}(i_q)+1$.
The half-open interval $[t_{base}^{start},t_{base}^{end})$ is then reused by both sequence and depth loops, ensuring index consistency.
For intuition, if $G=4$ and one query block contains rows $i_q=8,\ldots,15$, then
$t_{base}(i_q)\in\{2,3\}$, hence $t_{base}^{start}=2$ and $t_{base}^{end}=4$.

\myPara{Sequence attention loops}
The sequence phase contains two loops and both reuse the same accumulator states $(m,acc,o)$.
For fully visible blocks ($b_s<t_{base}^{start}$), we load $(\mathbf{K}_{[b_s]},\mathbf{V}_{[b_s]})$ from HBM to SRAM, compute
$S=\mathbf{Q}_{[b_q]}\mathbf{K}_{[b_s]}^\top/\sqrt{d}$, and call \textsc{OnlineSoftmaxUpdate}.
In this region, all keys are earlier than the current query base-time, so no causal mask is required.
For boundary blocks ($t_{base}^{start}\le b_s<t_{base}^{end}$), the same pipeline is used with grouped causal masking
$\lfloor i_q/G\rfloor\ge i_k$.
Hence, logits from multiple sequence blocks are accumulated into one online-softmax state without intermediate HBM materialization.
This is equivalent to processing a longer concatenated key sequence while keeping computation blockwise.

\myPara{Depth attention loop}
After sequence accumulation, the kernel enters the depth loop with flattened depth indices
$b_d\in[t_{base}^{start}L,t_{base}^{end}L)$.
The factor $L$ maps a base-time index to its contiguous depth span of length $L$.
For each depth block, $(\mathbf{K}^{\text{depth}}_{[b_d]},\mathbf{V}^{\text{depth}}_{[b_d]})$ is loaded from HBM to SRAM, and depth logits
$S_d=\mathbf{Q}_{[b_q]}(\mathbf{K}^{\text{depth}}_{[b_d]})^\top/\sqrt{d}$ are computed.
We then apply the depth mask
\[
mask(i_q,j_d)=\mathbf{1}\!\left[\left\lfloor\frac{i_q}{G}\right\rfloor=\left\lfloor\frac{j_d}{L}\right\rfloor\right]
:=
\begin{cases}
1, & j_d \in \left[L\left\lfloor\frac{i_q}{G}\right\rfloor,\ L\left(\left\lfloor\frac{i_q}{G}\right\rfloor+1\right)\right), \\
0, & \text{otherwise}.
\end{cases}
\]
which keeps only depth entries matched to the same base-time index as the query row.
The masked logits are then passed to \textsc{OnlineSoftmaxUpdate}, reusing the same $(m,acc,o)$ states as the sequence phase.
Finally, we normalize once on chip via $o\leftarrow o/acc$, write $\mathbf{O}_{[b_q]}$ back to HBM, and return $\mathbf{O}$ after all query blocks are processed.

\begin{table}[t]
    \caption{Efficiency comparison of hardware-efficient \name{} and FlashAttention-2 Triton kernels under ``forward\&backward'' setting.
    We report runtime (ms), depth utilization ($\eta_{\text{depth}}$), and relative extra time across three scaling settings.
    Here, $B$ denotes batch size, $d$ denotes head dimension, and $C$ denotes chunk size.
    We launch all experiments on A100 GPU with bfloat16 data type.}
    \label{tab:moda_hardware_efficiency}
    \centering
    \footnotesize
    \setlength{\tabcolsep}{4pt}
    \begin{tabular}{cccccccccc}
    \toprule
    No. & $T$ & $G$ & $H_q$ & $H_k$ & $L$ & FA2-triton (ms) & \name{}-triton (ms) & Depth Utilization ($\eta_{\text{depth}}$) & Extra Time Percentage \\
    \midrule
    \multicolumn{10}{c}{\bfseries Scaling Sequence Length $T$ ($B{=}1$, $d{=}64$, $C{=}64$)} \\
    \midrule
    (1)  & 4096  & 8  & 64  & 8 & 64  & 7.970    & 10.750   & 12.50\% & 25.86\% \\
    (2)  & 8192  & 8  & 64  & 8 & 64  & 28.700   & 35.427   & 12.50\% & 18.99\% \\
    (3)  & 16384 & 8  & 64  & 8 & 64  & 116.700  & 127.661  & 12.50\% & 8.59\% \\
    (4)  & 32768 & 8  & 64  & 8 & 64  & 459.854  & 480.914  & 12.50\% & 4.38\% \\
    (5)  & 65536 & 8  & 64  & 8 & 64  & 1831.668 & 1883.026 & 12.50\% & 2.73\% \\
    \midrule
    \multicolumn{10}{c}{\bfseries Scaling GQA Group Size $G$ ($B{=}1$, $d{=}64$, $C{=}64$)} \\
    \midrule
    (6)  & 16384 & 2  & 16  & 8 & 64  & 28.982  & 39.741  & 3.12\% & 27.07\% \\
    (7)  & 16384 & 4  & 32  & 8 & 64  & 58.071  & 68.939  & 6.25\% & 15.76\% \\
    (8)  & 16384 & 8  & 64  & 8 & 64  & 116.700 & 127.661 & 12.50\% & 8.59\% \\
    (9)  & 16384 & 16 & 128 & 8 & 64  & 233.700 & 244.900 & 25.00\% & 4.57\% \\
    (10) & 16384 & 32 & 256 & 8 & 64  & 467.107 & 480.767 & 50.00\% & 2.84\% \\
    \midrule
    \multicolumn{10}{c}{\bfseries Scaling Model Depth $L$ ($B{=}1$, $d{=}64$, $C{=}64$)} \\
    \midrule
    (11) & 16384 & 8 & 64 & 8 & 64  & 116.700 & 127.661 & 12.50\% & 8.59\% \\
    (12) & 16384 & 8 & 64 & 8 & 128 & 116.700 & 138.224 & 12.50\% & 15.57\% \\
    (13) & 16384 & 8 & 64 & 8 & 256 & 116.700 & 167.958 & 12.50\% & 30.52\% \\
    \bottomrule
    \end{tabular}
    \end{table}

\subsubsection{Efficiency Comparison}

Table~\ref{tab:moda_hardware_efficiency} reports end-to-end ``forward\&backward'' runtime of hardware-efficient \name{} against FlashAttention-2 Triton under controlled settings.
We sweep sequence length $T$, GQA group size $G$, and model depth $L$ while fixing the remaining factors in each block ($B{=}1$, $d{=}64$, $C{=}64$).
Besides raw runtime (ms), we also report depth utilization and the relative extra time percentage of \name{}.

When scaling sequence length, i.e., let $T$ increase from 4096 to 65536, with $G{=}8$, $L{=}64$, both kernels follow the expected growth trend, while the relative extra time percentage of \name{} consistently decreases from 25.86\% to 2.73\%.
This indicates that as sequence computation becomes dominant, the additional depth path is increasingly amortized.
When scaling group size $G$ from 2 to 32 at fixed $T{=}16384$, depth utilization rises from 3.12\% to 50.00\%, and the extra time percentage drops from 27.07\% to 2.84\%.

In contrast, when scaling model depth at fixed $T{=}16384$ and $G{=}8$, FlashAttention-2 runtime remains constant 116.700 ms, while \name{} runtime increases from 127.661 to 167.958 ms.
Accordingly, the extra time percentage rises from 8.59\% to 30.52\%, which is consistent with the fact that deeper depth streams introduce more depth-KV processing.
Overall, the results show that the proposed implementation has predictable linearly scaling behavior and remains efficient in long-sequence, high-utilization regimes.

\section{Experiment}
In this section, we demonstrate the expressivity and efficiency of the proposed \name{} through the experiments on Large Language Model (LLM).

\subsection{Experimental Setups}
\myPara{Model Architecture and Training Settings}
We conduct main experiments on language models of different sizes: 700M, and 1.5B.
Following the popular practice, we adopt group query attention (GQA)~\cite{ainslie2023gqa} for 700M and 1.5B models.
We train them on the 400B-token-subsets of OLMo2~\cite{olmo2024olmo2} dataset.
All models are trained in bfloat16 (bf16) precision.
The global batch size is set to 1024, and the context sequence length is set to 4096.
More detailed training configurations, such as learning rate schedule, AdamW~\cite{loshchilov2018decoupled} optimizer, etc., are following the OLMo2~\cite{olmo2024olmo2} implementation.

\myPara{Evaluation Details}
We evaluate the models on popular benchmarks, including PiQA~\cite{bisk2020piqa}, HellaSwag~\cite{zellers2019hellaswag}, WinoGrande~\cite{sakaguchi2021winogrande}, OpenBookQA~\cite{mihaylov2018openbookqa}, BoolQA~\cite{clark2019boolq}, SciQA~\cite{auer2023sciqa}, COPA~\cite{roemmele2011copa}, MMLU~\cite{hendrycks2020mmlu}, ARC-easy (ARC-E) and ARC-challenge (ARC-C)~\cite{clark2018arc}.
We further report the training perplexity (PPL), C4 validation perplexity (Val PPL), and per-domain validation perplexity on C4~\cite{raffel2020c4}, ICE~\cite{olmo2024olmo2}, m2d2-s2orc~\cite{lo2020s2orc}, Pile~\cite{gao2020pile}, Wiki-text~\cite{olmo2024olmo2}, and dolma~\cite{soldaini2024dolma} validation sets, which includes Books, Common Crawl, peS2o, Reddit, and Stack.

\subsection{Main Results}
\subsubsection{\name{} Variants}

\begin{table}[t]
    \caption{Performance of different mixture-of-depths attention (\name{}) variants on the training set, C4 validation set, and downstream benchmarks. 
    We train the 700M models on 400B tokens.
    For \textbf{\name} settings:
    \textbf{`Sequence KV'} means the each token only attends to the sequence keys/values, can be regarded as the vanilla attention mechanism.
    \textbf{`Depth KV'} means the each token attends to its depth keys/values.
    \textbf{`Extra FFN KV Proj.'} means further project the FFN's input $X$ to the depth keys/values, which are then used in subsequent attention operations.
    \textbf{`Extra Attn KV Proj.'} means set individual depth key/value projections rather than reuse the original key/value projections of sequence attention.
    The width $D$, GQA group size $G$, sequence length $T$ are set to 1024, 2, and 4096, respectively. 
    We further report the parameters and FLOPs of the models.}
    \label{tab:moda_variants}
    \footnotesize
    \begin{tabular}{l|ccccccc|ccc}
    \toprule
    \multirow{3}{*}{Model} & \multirow{3}{*}{Layer} & \multicolumn{4}{c}{Mixture-of-Depths  Attention (\name)}                                                                                                                                              & \multirow{3}{*}{\begin{tabular}[c]{@{}c@{}}Params\\ (M)\end{tabular}} & \multirow{3}{*}{\begin{tabular}[c]{@{}c@{}}FLOPs\\ (T)\end{tabular}} & \multirow{3}{*}{\begin{tabular}[c]{@{}c@{}}Train \\ PPL\end{tabular}} & \multirow{3}{*}{\begin{tabular}[c]{@{}c@{}}C4 Val \\ PPL\end{tabular}} & \multirow{3}{*}{\begin{tabular}[c]{@{}c@{}}Downstream\\ Average\end{tabular}} \\
                            &                        & \begin{tabular}[c]{@{}c@{}}Sequence \\ KV \end{tabular}                & \begin{tabular}[c]{@{}c@{}}Depth \\ KV \end{tabular}                   & \begin{tabular}[c]{@{}c@{}}Extra FFN \\ KV Proj.\end{tabular} & \begin{tabular}[c]{@{}c@{}}Extra Attn \\ KV Proj.\end{tabular} &                                                                       &                        &                                                                       &                                                                        &                                                                               \\
    \midrule
    \multicolumn{11}{c}{\bfseries Baseline Models} \\
    \midrule
    (1) OLMo2                  & 36                     & \ding{52} &                            &                                                                 &                                                                  & 669.0                                                                 & 8.01                & 14.49                                                                 & 18.59                                                                  & 56.93                                                                         \\
    (2) OLMo2                  & 38                     & \ding{52} &                            &                                                                 &                                                                  & 700.5                                                                 & 8.41                & 14.27                                                                 & 18.31                                                                  & 57.11                                                                         \\
    \midrule
    \multicolumn{11}{c}{\bfseries Our Models} \\
    \midrule
    (3) Ours                   & 36                     & \ding{52} & \ding{52} &                                                                 &                                                                  & 669.0                                                                 & 8.02                & 14.08                                                                 & 18.48                                                                  & 58.10                                                                         \\
    (4) Ours                   & 36                     & \ding{52} & \ding{52} & \ding{52}                                      &                                                                  & 705.7                                                                 & 8.33                & 13.90                                                                 & 18.21                                                                  & 58.87                                                                         \\
    (5) Ours                   & 36                     & \ding{52} & \ding{52} & \ding{52}                                      & \ding{52}                                       & 742.4                                                                 & 8.63                & 13.83                                                                 & 18.17                                                                  & 58.97 \\
    \bottomrule
    \end{tabular}
    \end{table}

We first compare the results of different mixture-of-depths attention (\name{}) variants on the 700M model size.
All models use a scheduler that warms up to a maximum learning rate of 3e-4 in 2k training steps, then decay to 3e-5 following the cosine schedule.
We present experimental results in Table~\ref{tab:moda_variants}.
To provide a fair comparison, we supplement the vanilla attention mechanism (OLMo2) as a baseline (row 1).
Because the extra FFN KV projection introduces additional parameters, we also report the more-parameter baseline (row 2) with two additional layers.
These methods introduce a comparable number of parameters/FLOPs than the proposed \name{} models.

From Table~\ref{tab:moda_variants}, we can observe that: 
\textbf{(i) Depth KV significantly improves performance}. 
Our method (row 3) keeps the same number of parameters as the baseline (row 1), but insert each token's depth KV into the attention computation.
Note that we directly reuse the preceding layer's sequence KV as the depth KV, which would not introduce additional projection parameters.
With only 0.12\% extra FLOPs, it improves 0.41 train PPL, 0.11 C4 validation PPL, and 1.17 downstream averaged metrics (row 1 vs. row 3).
\textbf{(ii) FFN layers' depth KV matters}.
Experiment in row 3 only considers treat the preceding attention layers' KV as the depth KV, which ignores the FFN layers.
We further add additional KV projections to enhance the original FFN , which projects the FFN's input $X$ to its corresponding depth keys/values.
Comparing row 3 and row 4, we can observe that incorporating KV from FFN improves 0.18 train PPL, 0.27 C4 validation PPL, and 0.77 downstream averaged metrics.
While comparing row 4 with more-parameter baseline (row 2), it improves 0.37 train PPL, 0.10 C4 validation PPL, and 1.76 downstream averaged metrics.
Notably, row 4 has similar number of parameters/FLOPs as row 2, but achieves better performance, which demonstrates that FFN's depth information also contributes to the mixture-of-depths attention (\name).
\textbf{(iii) Extra Attn KV Projection is overly saturated}.
Based on the row 4, we further introduce additional depth KV projection, which specifically projects the attention layers' input $X$ to the depth keys/values.
Comparing row 4 and row 5, we can observe that incorporating Extra Attn KV Projection only improves 0.07 train PPL, 0.04 C4 validation PPL, and 0.10 downstream averaged metrics.
However, this modification introduces a non-trivial overhead (from 705.7M to 742.4M parameters and from 8.33T to 8.63T FLOPs), indicating that the additional attention-side depth projection is close to saturation.

Overall, these experiments reveal a clear design principle for \name{}: injecting depth information is effective, but the gains are highly sensitive to where additional projections are introduced.
In particular, reusing attention-side depth KV already provides strong improvements at almost no cost, while adding FFN-side depth KV yields the best accuracy-efficiency trade-off.
By contrast, introducing extra attention KV projections brings only marginal gains with noticeable parameter/FLOPs overhead.
Therefore, we adopt the setting in row 4 as the default \name{} variant in the following scaling up experiments (Section \ref{subsec:scaling_moda}).

\begin{table}[t]
    \caption{Performance of the proposed \textbf{\name} models with varying model sizes on the downstream benchmarks.
    We train the 700M and 1.5B models on the 400B tokens of OLMo2 dataset.
    The width $D$, GQA group size $G$, sequence length $T$ are set to 1024, 2, and 4096, respectively.
    We mark the best performance with the \textbf{bold} font.
    }
    \label{tab:downstream_results}
    \centering
    \footnotesize
    \begin{tabular}{l|cccccccccccc} 
    \toprule
    Model                 & PIQA & \makecell{Hella-\\Swag} & \makecell{Wino-\\Grade} & \makecell{OpenBook-\\QA} & BoolQA & SciQ & ARC-E & ARC-C & COPA & MMLU & Average  \\
    \midrule
    \multicolumn{12}{c}{\bfseries 700M Models} \\
    \midrule
    (1) OLMo2 & \textbf{73.72} & 58.77 & 55.33 & 35.60 & 56.24 & 89.50 & 66.84 & 33.44 & 77.00 & 24.69 & 57.11 \\
    (2) Ours & 73.39 & \textbf{59.19} & \textbf{60.22} & \textbf{37.20} & \textbf{59.33} & \textbf{89.60} & \textbf{67.37} & \textbf{34.78} & \textbf{82.00} & \textbf{25.61} & \textbf{58.87} \\
    \midrule
    \multicolumn{12}{c}{\bfseries 1.5B Models} \\
    \midrule
    (3) OLMo2 & 76.55 & 65.86 & 63.22 & 38.80 & 63.61 & 90.60 & \textbf{72.98} & 42.47 & 81.00 & 27.73 & 62.28 \\
    (4) Ours & \textbf{76.82} & \textbf{66.24} & \textbf{65.59} & \textbf{41.60} & \textbf{67.34} & \textbf{92.10} & 72.81 & \textbf{46.82} & \textbf{85.00} & \textbf{29.59} & \textbf{64.39} \\
    \bottomrule
    \end{tabular}
    \end{table}

\begin{table}[t]
    \caption{Per-domain validation perplexity of the proposed \textbf{\name} models with varying model sizes.
    We train the 700M and 1.5B models on the 400B tokens of OLMo2 dataset.
    The width $D$, GQA group size $G$, sequence length $T$ are set to 1024, 2, and 4096, respectively.
    Lower perplexity indicates better performance and is marked with the \textbf{bold} font.
    }
    \label{tab:domain_ppl_results}
    \centering
    \footnotesize
    \begin{tabular}{l|ccccccccccc}
    \toprule
    Model & C4 & ICE & m2d2-s2orc & Pile & Wiki-text & Books & CC & peS2o & Reddit & Stack & Average \\
    \midrule
    \multicolumn{12}{c}{\bfseries 700M Models} \\
    \midrule
    (1) OLMo2 & 18.32 & 17.43 & 24.37 & 9.53 & 12.26 & 16.78 & 20.53 & 9.17 & 23.84 & 3.93 & 15.61 \\
    (2) Ours & \textbf{18.29} & \textbf{17.24} & \textbf{23.64} & \textbf{9.48} & \textbf{12.06} & \textbf{16.58} & \textbf{20.52} & \textbf{9.14} & \textbf{23.75} & \textbf{3.90} & \textbf{15.46} \\
    \midrule
    \multicolumn{12}{c}{\bfseries 1.5B Models} \\
    \midrule
    (3) OLMo2 & 16.16 & 15.37 & 21.10 & 8.45 & 10.41 & 14.19 & 18.13 & 8.19 & 21.21 & 3.57 & 13.67 \\
    (4) Ours & \textbf{15.97} & \textbf{15.08} & \textbf{20.92} & \textbf{8.33} & \textbf{10.16} & \textbf{13.95} & \textbf{17.88} & \textbf{8.09} & \textbf{20.85} & \textbf{3.52} & \textbf{13.47} \\
    \bottomrule
    \end{tabular}
\end{table}

\subsubsection{Scaling \name{} with Model Size}
\label{subsec:scaling_moda}

We study whether the gains of \name{} persist when scaling model size from 700M to 1.5B under the same training budget of 400B tokens.
We report downstream benchmark results in Table~\ref{tab:downstream_results} and domain-level validation perplexity in Table~\ref{tab:domain_ppl_results}.
From these two tables, we can observe that:
\textbf{(i) \name{} provides stable average gains on downstream benchmarks across model scales.}
For 700M models, row 1 v.s. row 2 in Table~\ref{tab:downstream_results} improves the average from 57.11 to 58.87, which is +1.76.
For 1.5B models, row 3 v.s. row 4 improves the average from 62.28 to 64.39, which is +2.11.
\textbf{(ii) Downstream gains are broadly observed across commonsense, reasoning, and broad-knowledge tasks.}
On commonsense and causal discrimination tasks, the gains on HellaSwag, WinoGrande, and COPA are +0.42, +4.89, and +5.00 at 700M, and +0.38, +2.37, and +4.00 at 1.5B.
On science-oriented and harder reasoning tasks, the gains on OpenBookQA, ARC-C, and SciQ are +1.60, +1.34, and +0.10 at 700M, and +2.80, +4.35, and +1.50 at 1.5B.
We also observe consistent gains on broad-knowledge benchmarks, including BoolQ with +3.09 and +3.73, and MMLU with +0.92 and +1.86, for 700M and 1.5B, respectively.
\textbf{(iii) Validation perplexity gains are broad and consistent across domains.}
In Table~\ref{tab:domain_ppl_results}, row 1 v.s. row 2 at 700M lowers average PPL from 15.61 to 15.46 and improves all ten domains.
The largest 700M reduction appears on m2d2-s2orc, where PPL decreases from 24.37 to 23.64.
At 1.5B, row 3 v.s. row 4 lowers average PPL from 13.67 to 13.47 and also improves all ten domains.
Notable 1.5B reductions appear on Reddit from 21.21 to 20.85, on ICE from 15.37 to 15.08, and on Wiki-text from 10.41 to 10.16.

Overall, the two tables provide consistent evidence from complementary evaluation views.
Table~\ref{tab:downstream_results} shows improvements on end-task performance, while Table~\ref{tab:domain_ppl_results} shows improved language modeling quality across diverse domains.

\begin{table}[]
    \caption{Layer-number analysis of \name{} under deeper (48-layer) and shallower (24-layer) model settings.
    We compare vanilla attention (OLMo2) and \name{} variants with different \name{} choices, under both pre-norm and post-norm configurations.
    Models are trained with the same data recipe, and we report parameter count, FLOPs, and FineWeb-Edu validation loss.
    Across both depth regimes, introducing Depth KV consistently improves validation loss, and adding Extra FFN KV Projection yields further gains at moderate compute overhead.}
    \footnotesize
    \begin{tabular}{l|ccccccc|c}
    \toprule
    \multirow{2}{*}{Model} & \multirow{2}{*}{Layer} & \multirow{2}{*}{Norm} & \multicolumn{3}{c}{Mixture-of-Depths Attention (MoDA)}                                                                                                                      & \multirow{2}{*}{\begin{tabular}[c]{@{}c@{}}Params\\ (M)\end{tabular}} & \multirow{2}{*}{\begin{tabular}[c]{@{}c@{}}FLOPs\\ (G)\end{tabular}} & \multirow{2}{*}{\begin{tabular}[c]{@{}c@{}}FineWeb-Edu\\ Val Loss\end{tabular}} \\
                           &                        &                       & Sequence KV & Depth KV & Extra FFN KV Proj. &                                                                       &                                                                      &                                                                                 \\
    \midrule
                           \multicolumn{9}{c}{\bfseries Experiments with Deeper Models (48 Layers)} \\        
    \midrule
    (1) OLMo2                  & 48                     & prenorm               & \ding{52}                            &                                                    &                                                               & 123.38                                                                & 136.61                                                          & 3.3800                                                                            \\
    (2) OLMo2                  & 48                     & postnorm              & \ding{52}                            &                                                    &                                                               & 123.38                                                                & 136.61                                                          & 3.4062                                                                          \\
    (3) Ours                   & 48                     & prenorm               & \ding{52}                            & \ding{52}                         &                                                               & 123.38                                                                & 137.89                                                          & 3.3759                                                                          \\
    (4) Ours                   & 48                     & postnorm              & \ding{52}                            & \ding{52}                         &                                                               & 123.38                                                                & 137.89                                                          & 3.3653                                                                          \\
    (5) Ours                   & 48                     & prenorm               & \ding{52}                            & \ding{52}                         & \ding{52}                                    & 128.11                                                                & 144.00                                                          & 3.3656                                                                          \\
    (6) Ours                   & 48                     & postnorm              & \ding{52}                            & \ding{52}                         & \ding{52}                                    & 128.11                                                                & 144.00                                                          & 3.3484                                                                          \\
    \midrule
    \multicolumn{9}{c}{\bfseries Experiments with Shallower Models (24 Layers)} \\
    \midrule
    (7) OLMo2                  & 24                     & postnorm              & \ding{52}                            &                                                    &                                                               & 71.35                                                                 & 78.19                                                          & 3.4740                                                                           \\
    (8) Ours                   & 24                     & postnorm              & \ding{52}                            & \ding{52}                         &                                                               & 71.35                                                                 & 78.51                                                         & 3.4537                                                                          \\
    (9) Ours                   & 24                     & postnorm              & \ding{52}                            & \ding{52}                         & \ding{52}                                    & 73.72                                                                 & 81.24                                                          & 3.4338 \\
    \bottomrule
    \end{tabular}
    \end{table}

\subsection{Analysis}

\subsubsection{Analyzing \name{} with Layer Number}

To study whether \name{} remains effective under different depth budgets, we conduct layer-number experiments on small models using the FineWeb-Edu data pipeline.
We reserve an additional held-out split from FineWeb-Edu for validation and report validation loss for all settings.
Specifically, we evaluate both deeper models (48 layers) and shallower models (24 layers), and compare vanilla attention with \name{} variants under pre-norm/post-norm configurations.
For all runs in this subsection, the model width is 384, the number of query heads is 6, and the number of key/value heads is 2.

From the layer-number results, we observe that:
\textbf{(i) Depth KV consistently improves validation loss across different layer numbers.}
For 48-layer models, adding Depth KV reduces loss from 3.3800 to 3.3759 in pre-norm setting (row 1 vs. row 3), and from 3.4062 to 3.3653 in post-norm setting (row 2 vs. row 4).
For 24-layer models, adding Depth KV also reduces loss from 3.4740 to 3.4537 (row 7 vs. row 8).
\textbf{(ii) In deeper models, post-norm benefits more from Depth KV than pre-norm.}
At 48 layers, row 2 vs. row 4 gives a loss reduction of 0.0409 in post-norm, while row 1 vs. row 3 gives 0.0041 in pre-norm.
This indicates that Depth KV has stronger optimization impact in the post-norm configuration for deeper stacks.
\textbf{(iii) Extra FFN KV Projection provides additional gains on top of Depth KV.}
For 48-layer models, adding Extra FFN KV Projection further reduces loss from 3.3759 to 3.3656 in pre-norm (row 3 vs. row 5), and from 3.3653 to 3.3484 in post-norm (row 4 vs. row 6).
For 24-layer models, it further reduces loss from 3.4537 to 3.4338 (row 8 vs. row 9).
Overall, these results show that \name{} remains effective under layer scaling, and FFN-side depth information brings additional gains when compute budget allows.

\begin{figure}[htp]
    \centering
    \includegraphics[width=1\textwidth]{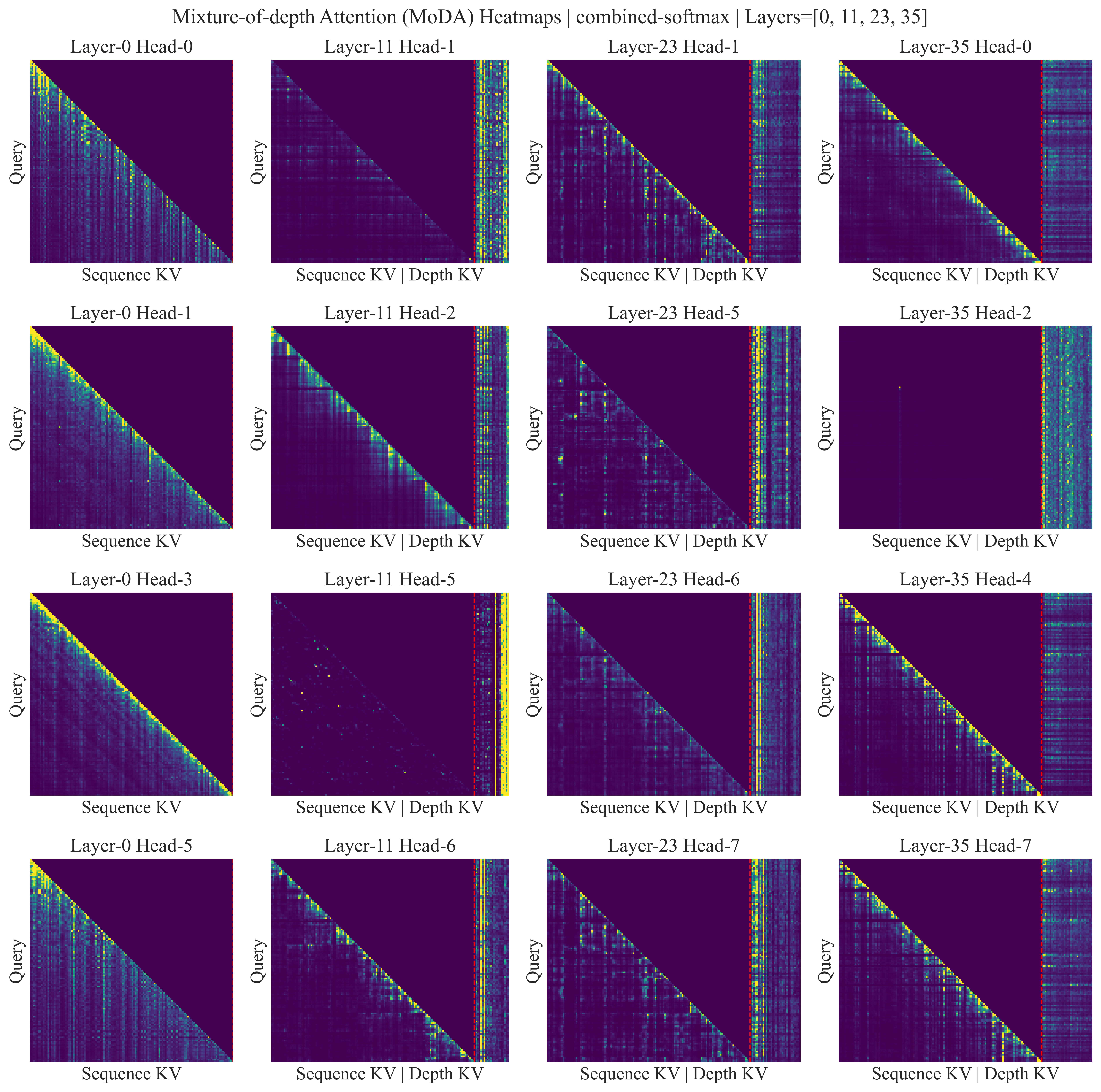}
    \caption{Mixture-of-Depths Attention (\name{}) heatmaps with the combined-softmax formulation.
    Columns correspond to uniformly sampled layers $\{0, 11, 23, 35\}$, and rows correspond to randomly selected heads in each layer.
    The first column shows attention over sequence KV only, while the other columns show concatenated \textit{Sequence KV | Depth KV}; the red dashed line marks the boundary between the two KV blocks.
    Across layers and heads, substantial attention mass is consistently assigned to the depth-KV block, indicating that \name{} effectively leverages depth information in addition to standard sequence attention.}
    \label{fig:attn_vis}
\end{figure}

\subsubsection{Analyzing \name{} with Attention Visualization}
To better understand how \name{} changes token interactions, we visualize attention heatmaps for the \textbf{700M} model trained on \textbf{400B tokens} (Fig.~\ref{fig:attn_vis}).
Under the combined-softmax formulation, each query attends over the concatenated \textit{Sequence KV | Depth KV} space (red dashed line indicates the boundary).
Notably, the depth-KV part contains both attention KV and FFN KV.

From the heatmaps, we observe non-trivial and persistent attention mass on the depth-KV block, especially in middle and late layers.
This indicates that the model actively retrieves cross-layer depth information instead of relying only on sequence-local context.
We also find a complementary pattern: heads with sharper diagonal sequence attention still allocate part of probability to depth slots, while broader heads tend to rely more heavily on depth-KV entries.

Another important observation is that \name{} exhibits attention patterns that differ from the typical \emph{attention sink} behavior observed in the visualized heads.
Rather than collapsing a large fraction of probability mass onto a few fixed sink positions, the attention in these heads appears to be distributed more broadly across sequence and depth slots, including slots that may be relevant to the task.

This qualitative difference suggests that \name{} may alter how attention mass is allocated in long-context settings.
In particular, the visualization indicates that some probability mass is redistributed away from fixed sink positions toward sequence/depth locations that potentially carry useful information.
While these patterns are intriguing, their precise functional role remains unclear and warrants further investigation.

Overall, the visualization is consistent with the core intuition of \name{}:
depth information can serve as a complementary retrieval channel to standard sequence attention.
At the same time, the changed attention-sink pattern may point to additional mechanisms or insights beyond the original design motivation, which should be studied more carefully in future work.

\subsubsection{Analyzing \name{} with Efficiency}

To quantify the practical efficiency contribution of each kernel design, we perform an incremental ablation and report end-to-end ``forward\&backward'' runtime in Table~\ref{tab:kernel_efficiency_ablation}.
All experiments are conducted on a single A100 GPU with bfloat16 under fixed setting $B{=}1,\,T{=}1024,\,G{=}8,\,H_q{=}64,\,H_k{=}8,\,d{=}64,\,L{=}64,\,C{=}64$.
Noting that naive PyTorch implementation is not optimized for efficiency, we only report the comparison under a short sequence length, i.e., $T{=}1024$.

\begin{table}[t]
    \caption{Ablation of the proposed kernel implementation strategies under a fixed configuration.
    Each row incrementally enables one optimization component: (1) naive PyTorch baseline, (2) Flash-compatible depth-KV layout, (3) Flash-compatible \& chunk-aware depth-KV layout, and (4) Flash-compatible \& chunk-aware \& group-aware indexing.
    We report end-to-end ``forward\&backward'' runtime in milliseconds (ms), where lower is better and the best performance is marked with the \textbf{bold} font.
    Experiments are run on a single A100 GPU with bfloat16 under fixed setting $B{=}1,\,T{=}1024,\,G{=}8,\, H_q{=}64,\, H_k{=}8,\, d{=}64,\, L{=}64,\, C{=}64$.}
    \label{tab:kernel_efficiency_ablation}
    \centering
    \footnotesize
    \begin{tabular}{lccccc}
    \toprule
    No. & Naive Torch & Flash-Compatible & Chunk-Aware & Group-Aware & Time (ms) \\
    \midrule
    \multicolumn{6}{c}{\bfseries $B{=}1,\,T{=}1024,\,G{=}8,\, H_q{=}64,\, H_k{=}8,\, d{=}64,\, L{=}64,\, C{=}64$} \\
    \midrule
    (1) & \ding{52} &           &           &           & 2128.900 \\
    \midrule
    (2) &           & \ding{52} &           &           & 13.102   \\
    (3) &           & \ding{52} & \ding{52} &           & 6.286    \\
    (4) &           & \ding{52} & \ding{52} & \ding{52} & \textbf{1.460} \\
    \bottomrule
    \end{tabular}
\end{table}

From Table~\ref{tab:kernel_efficiency_ablation}, we observe that:
\textbf{(i) Flash-compatible depth-KV layout already provides orders-of-magnitude acceleration over naive implementation.}
Row 1 vs. row 2 reduces runtime from 2128.900 ms to 13.102 ms, i.e., about 162.5$\times$ faster.
\textbf{(ii) Chunk-aware depth-KV layout further improves efficiency by reducing memory-access overhead.}
On top of Flash compatibility, row 2 vs. row 3 lowers runtime from 13.102 ms to 6.286 ms, corresponding to a 52.0\% reduction.
\textbf{(iii) Group-aware indexing is essential for fully exploiting group reusing mechanism.}
Adding group-aware indexing (row 3 vs. row 4) further reduces runtime from 6.286 ms to 1.460 ms, giving an additional 4.31$\times$ speedup.
Overall, combining all three optimizations yields the best runtime and achieves about 1458$\times$ end-to-end speedup over the naive PyTorch baseline (row 1 vs. row 4).

\section{Conclusion}
In this paper, we present \name{}, a unified depth-aware attention mechanism for LLM to improve depth-wise information aggregating and mitigate depth-efficiency gaps from optimization difficulty and information dilution. 
We further develop a hardware-aware fused kernel with unified online-softmax states, chunk-aware depth-KV layout, and group-aware indexing to maintain efficient long-context execution. 
Experiments on 700M and 1.5B models trained with the OLMo2 recipe show consistent gains in perplexity and downstream performance under modest overhead. 
These results suggest that explicit retrieval of historical depth information is a practical and effective primitive for scaling Transformer depth.
We will release the full implementation of \name{}, and we hope it will serve as a foundation for building stronger large language models in the open-source community.
Beyond language modeling, \name{} is architecture-agnostic and can be readily integrated into multimodal intelligence, visual understanding, and world models, where Transformers are increasingly adopted.
We believe that principled depth-aware information aggregating will bring broad and lasting benefits across these diverse domains.

\section{Discussion}

\subsection{Scaling MoDA for Industrial Training via Advanced CUDA Engineering}
Although the current hardware-aware \name{} kernel already achieves competitive efficiency against FlashAttention-2, 
it is not yet the endpoint for industrial-scale training, e.g., trillion-parameter models. 
In large production runs, additional CUDA engineering remains critical, including improved memory scheduling, deeper computation pipelining, and tighter overlap between fused attention kernels and distributed communication. 
These optimizations do not change the algorithmic behavior of \name{}, but can further reduce memory stalls and kernel-launch overhead, improve end-to-end throughput, and increase cluster-level training efficiency. 
Therefore, we view future CUDA optimization as an important direction for turning \name{} from an efficient research operator into a robust primitive for industrial LLM training.

\subsection{Mitigating Memory Bottlenecks with Bounded Depth-KV Slot Caching
}
When scaling to very deep networks, caching all depth-KV states from all historical layers introduces substantial memory and bandwidth overhead.
The cost grows linearly with depth, and can become the dominant bottleneck in long-context training and serving.
As a result, full depth-KV caching is increasingly hard to sustain at industrial scale.

A practical direction is to use a fixed-size Depth KV slot buffer.
Instead of storing all depth-KV entries, each query only attends to a bounded set of slots.
The slot budget is fixed to $S$, where $S \ll L$, and the system dynamically decides which depth-KV entries are kept.
Two policies are natural.
One is dynamic selection, which scores candidate depth-KV entries by utility and keeps the top-$S$ entries.
The other is a sliding-window policy, which keeps the most recent depth-KV entries and evicts older ones.
A hybrid design can also be used, where part of the slots are reserved for recency and the rest for high-score global memories.

This design changes the effective depth memory from an unbounded cache to a bounded cache.
The memory and bandwidth terms move from depth-dependent scaling to slot-dependent scaling.
It also provides a stable tensor shape for fused kernel implementation.
In practice, the key challenge is the quality of slot assignment.
Future work should study how to train the selection policy jointly with \name{}, and how to balance quality, latency, and hardware efficiency under a fixed slot budget.

\bibliographystyle{plainnat}
\bibliography{main}

\end{document}